\documentclass[letterpaper,10pt,conference]{ieeeconf}
\IEEEoverridecommandlockouts
\usepackage[T1]{fontenc}
\usepackage{float}
\usepackage{amsfonts}
\usepackage{amsthm}
\usepackage{pifont}
\usepackage{xcolor}
\usepackage{soul}

\newtheorem{definition}{Definition}
\newtheorem{proposition}{Proposition}

\newtheorem{problem}{Problem}
\usepackage{glossaries}
\usepackage{listings}
\usepackage{amsmath}
\usepackage{amssymb}
\usepackage{subfig}
\usepackage{graphicx}
\usepackage{stix}
\usepackage{tikz}
\usepackage{etoolbox}
\usetikzlibrary{shapes,arrows}
\usepackage{verbatim}
\usepackage[hidelinks]{hyperref}

\renewcommand{\baselinestretch}{.985}

\newcommand{\deleted}[1]{}

\begin{document}
\title{\bf 
H-ModQuad: 
Modular Multi-Rotors with 4, 5, and 6 Controllable DOF
}
\author{Jiawei Xu, Diego S. D'Antonio, and David Saldaña
\thanks{J. Xu, D. D'Antonio, and D. Salda\~{n}a are with the Autonomous and Intelligent Robotics Laboratory (AIRLab), Lehigh University, PA, USA:$\{$\texttt{jix519, diego.s.dantonio, saldana\}@lehigh.edu}}
}  
\maketitle
\begin{abstract}
Traditional aerial vehicles are usually custom-designed for specific tasks. Although they offer an efficient solution, they are not always able to adapt to changes in the task specification, e.g., increasing the payload.
This applies to quadrotors, having a maximum payload and 
only four controllable degrees of freedom, limiting their adaptability to the task's variations.
We propose a versatile modular robotic system that can increase its payload and degrees of freedom by assembling heterogeneous modules; we call it H-ModQuad.
It consists of cuboid modules propelled by quadrotors with tilted propellers that can generate forces in different directions. By connecting different types of modules, an H-ModQuad can increase its controllable degrees of freedom from 4 to 5 and 6.
We model the general structure and propose three controllers, one for each number of controllable degrees of freedom.
We extend the concept of the actuation ellipsoid to find the best reference orientation that can maximize the performance of the structure.
Our approach is validated with experiments using actual robots, showing the independence of the translation and orientation of a structure.
%

\end{abstract}

\section{Introduction}

During the last decade, unmanned aerial vehicles (UAV) have significantly impacted our society. They offer low-cost solutions for a large variety of applications in comparison with the previous technology in aerial robots. 
However, one of the problems of aerial vehicles is their lack of versatility.
For instance, in object-transportation, the robot hardware depends on the weight and dimensions of the object.
On one hand, large vehicles can carry many types of objects in contrast to  small vehicles that can only carry lightweight payloads.
On the other hand, large vehicles have a high cost and motion limitations in small spaces.
In addition to strength, other characteristics are essential for aerial vehicles. A quadrotor has four controllable degrees of freedom (DOF) independently of its size.
Linear controllers \cite{5569026, 1302409} focus on achieving stability around the nominal hover state, defining the desired location and yaw orientation.
In \cite{5717652, 5980409}, the authors present a geometric controller in $\mathsf{SO(3)}$ such that the quadrotor converges to the desired attitude that can differ from the nominal hover state, which allows the quadrotor to perform aggressive maneuvers.
However, changing the internal controller, adding more vertical propellers to a UAV, \cite{alaimo2013mathematical}, or connecting multiple modular quadrotors \cite{8461014}, without changing the quadrotor design, does not increase the number of controllable DOF. The fact that all propellers are vertical and pointing in the same direction makes it impossible to drive the UAV horizontally without tilting.





An approach to dealing with this problem is to increase the number of propellers in a multirotor system but tilting them in different orientations. In \cite{7139759} is introduced the design and control strategy for a hexarotor with tilting propellers; using 6 rotors, the UAV can control its 6 DOF,  but the actuation range is limited. Another solution is to actively tilt the propellers in a quadrotor. In \cite{6225129} is proposed a control design for a quadrotor with controllable tilting propellers. In \cite{7759271}, the authors show a hybrid design where a hexarotor can transit from under-actuated to fully-actuated by using an additional servomotor to tilt the rotors synchronously. Those designs achieve the full actuation but the maximum force is still limited.
Although the work in \cite{franchi2018full} provides a universal control strategy for both under-actuated and fully-actuated UAVs,
the design of the UAV will be specific for a specific task, coming back to the versatility problem.

Modular aerial systems offer a versatile and scalable way to increase their strength \cite{8461014,duffy2015lift,Mellinger2012CooperativeQuadrotors, Oung2011TheArray}, and capabilities \cite{yang-icra19-aerialskeleton, zhaoIJRR2019, modquadgripper}. However, homogeneous modules can increase the total force, but not the controllability of the UAV. More modules only add more redundancy to the system without increasing the number of controllable DOF. 
We want to design an aerial system that can adapt to the problem without changing the design of the robot. If the task is about transporting a pizza or a piano, the difference would be in the number of modules that compose the aerial vehicle.
Even if the number of controllable DOF remain the same, the direction of the total force change depending on the module configuration.
Some applications require a high vertical force, e.g., lifting an object, and other applications require an accurate horizontal force, e.g., drilling a hole on a wall.

\begin{figure}[t!]
    \includegraphics[width=8.6cm]{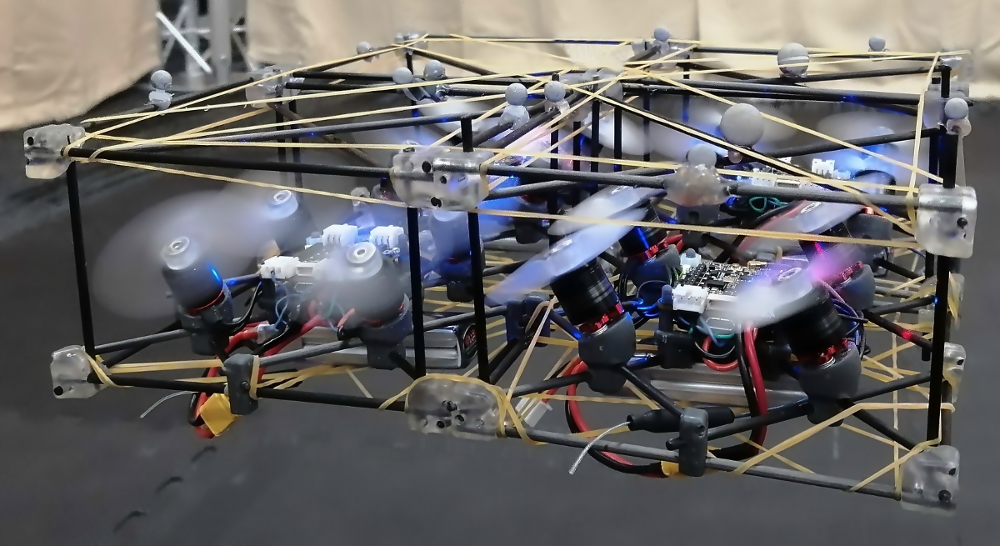}
    \caption{Four heterogeneous balanced modules forming an H-ModQuad structure. Each module has all its propellers pointing in the same direction, but different modules can have different propeller orientations.}
    \label{fig:titlePic}
\end{figure}

The main contribution of this paper is threefold.
First, we propose a modular UAV composed of individual heterogeneous modules that can fly by themselves, but together can increase their strength and controllable DOF from 4 to 5 and~6. 
Second, we extend the concept of the actuation ellipsoid to find the best reference orientation that can maximize the performance of the structure.
Third, we propose three controllers, one for each number of controllable degrees of freedom.

\section{Problem statement}
\noindent
Our work is focused on modular aerial robots composed of independent modules.
\begin{definition}[Module]
A \emph{module} is a UAV that can move by itself in a three-dimensional environment.  
It is composed of a quadrotor within a cuboid frame, a square base with dimensions  $l\times l$, and a height  $h$. 
The propellers do not necessarily have to be vertical with respect to the base of the module, and their orientations define the actuation properties of the module. 
\end{definition}
Each module has a mass $m$ and inertia tensor $\mathbf{I}_M$.
A module can dock to another by horizontally aligning the vertical faces of both,
creating a rigid connection. Performing multiple docking actions, we assemble a larger structure.
\begin{definition}[Structure]
A \emph{structure} is a set of $n\geq1$ rigidly connected modules
that behave as a single multi-rotor vehicle. These modules are
horizontally connected by docking along the sides, so the
resulting shape has the same height~$h$.
Its inertia tensor is denoted by $\mathbf{I}_S$.
\end{definition}

\begin{figure}[t]
\centering
    \includegraphics[width=0.6\linewidth]{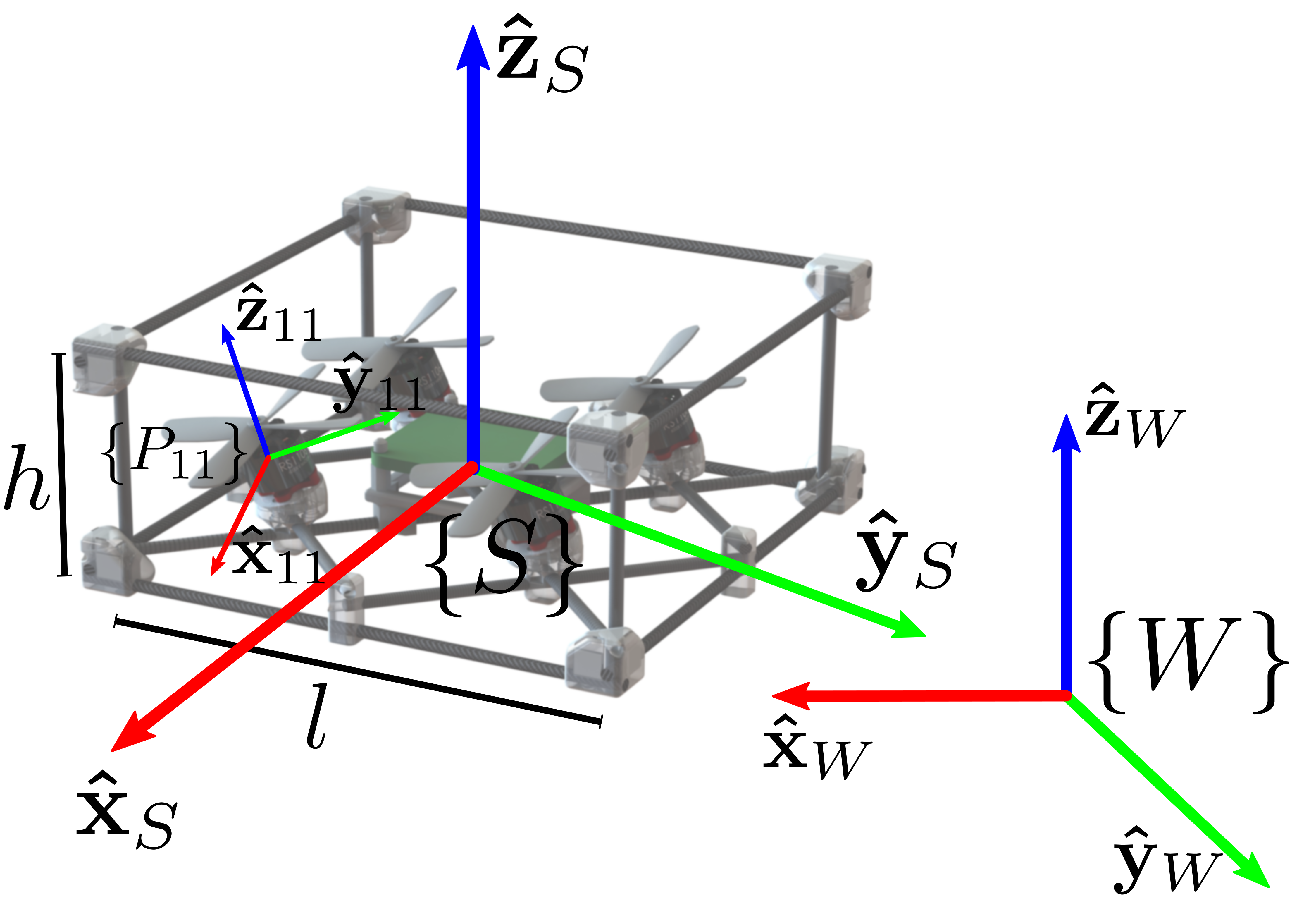}
    \caption{A module with its coordinate frames, and dimensions. Note that this module composes a structure, thus $\{S\}$ aligns with $\{M_1\}$. In Section \ref{sec:control}, we will define $F$-frame. In this example, the $F$-frame of the structure aligns with $\{P_{11}\}$.}
    \label{fig:oneUAV}
\end{figure}

We denote the 
standard basis in $\mathbb{R}^3$ by $\mathbf{\hat e}_1=[1,0,0]^\top,\: \mathbf{\hat e}_2=[0,1,0]^\top,$ and $\mathbf{\hat e}_3=[0,0,1]^\top$. 
The world coordinate frame, $\{W\}$, 
is fixed, and its $z$-axis points upwards. 
The $i$-th module in the structure has a module reference frame, $\{M_i\}$, with its origin in the module's center of mass. 
The four propellers are in a square configuration and located on the $xy$-plane. Each propeller has a propeller frame.
The $x$-axis points towards the front of the module, and the $y$-axis to its left. In a traditional quadrotor, the propellers are oriented vertically, parallel to the $z$-axis of the quadrotor. In our case, the propellers can be pointing in different directions. The orientation of the $j$-th propeller frame, $\{P_{ij}\}$, in the module frame, $\{M_i\}$, is specified by the rotation matrix ${}^{M_{i}}\mathbf{R}_{j}\in \mathsf{SO}(3)$. The associated coordinate frames of a module are illustrated in Fig. \ref{fig:oneUAV}. The structure frame, denoted by $\{S\}$, has its origin in its center of mass, and its  $x$-, $y$- and $z$-axes are parallel to the $x$-, $y$- and $z$-axes of the first module respectively.
The location and the orientation of $\{S\}$ in the world frame $\{W\}$ is specified by the vector $\mathbf{r}\in \mathbb{R}^3$ and the rotation matrix ${}^{W}\mathbf{R}_{S}\in \mathsf{SO}(3)$.
The angular velocity of each propeller $j\in\{1,...,4\}$ in a module $i$ 
generates a thrust $f_{ij}\in[0, f_{max}]$,
and a torque by the air drag. The force and torque vectors in $\{M_i\}$, as a function of the thrust $f_{ij}$, are 
\begin{equation*}
\mathbf{f}_{ij}=f_{ij}\:  {{}^{M_i}\mathbf{R}_{j} \mathbf{\hat e}_3}  
\text{,     and      }    
\boldsymbol\tau_{ij}=f_{ij}\: {(-1)^{j+1}\frac{k_m}{k_f}\:   {}^{M_i}\mathbf{R}_{j}\mathbf{\hat e}_3}
,
\end{equation*}
where $k_f$ and $k_m$ are coefficients that can be obtained experimentally.
%
%
The structure generates a total force $\mathbf{f}$ and torque $\boldsymbol\tau$ in the structure frame~$\{S\}$ which can be expressed as the summation of all local forces and torques
\begin{eqnarray}
    \mathbf{f}&=&\sum_{ij}{}^{S}\mathbf{R}_{M_i}\mathbf{f}_{ij}.
    \label{eq:fstructure}\\
    \boldsymbol\tau&=&\sum_{ij}\mathbf{p}_{ij}\times{}^{S}\mathbf{R}_{M_i}\mathbf{f}_{ij}+{}^{S}\mathbf{R}_{M_i}\boldsymbol\tau_{ij},
    \label{eq:taustructure}
\end{eqnarray}
where $\mathbf{p}_{ij}\in\mathbb{R}^3$ is the position of each propeller in $\{S\}$, and ${}^{S}\mathbf{R}_{M_i}$ is the orientation of the $i$-th module in $\{S\}$.

We use Newton-Euler's equation to describe the dynamics of the structure
\begin{eqnarray}
    nm\,\mathbf{\Ddot{r}}-nm\,g\mathbf{\hat e}_3 &=& {}^W\mathbf{R}_S\mathbf{f}\label{eq:newton},\\
    \mathbf{I}_S\Dot{\boldsymbol\omega}+\boldsymbol\omega\times \mathbf{I}_S\boldsymbol\omega &=& \boldsymbol\tau\label{eq:euler},
\label{eq:dynamics}
\end{eqnarray}
where 
$\mathbf{\Ddot{r}}$ is the linear acceleration of the structure, $\boldsymbol\omega, \boldsymbol{\dot{\omega}}$ are the angular velocity and acceleration of the structure, respectively, and $g$ is the gravity constant. 
Letting $\mathbf{u} = [f_{11}, f_{12}, ...,  f_{n4}]^{\top}$ be the input vector, we can rewrite \eqref{eq:fstructure} and \eqref{eq:taustructure} in a compact form 
\begin{equation}
    \begin{bmatrix}
                \mathbf{f}^\top &
                \boldsymbol\tau^\top
    \end{bmatrix}^\top
    =
    \mathbf{A}\mathbf{u}, \text{ where}
    \label{eq:MAu}
\end{equation}
\begin{equation}
    \centering
    \mathbf{A} = \begin{bmatrix}
    {}^{S}\mathbf{R}_{M_1}\mathbf{\hat e}_3& \cdots \\
    \mathbf{p}_{11}\times{}^S\mathbf{R}_{M_1}\mathbf{\hat e}_3+{}^{S}\mathbf{R}_{M_1}(-1)^{1+1}\frac{k_m}{k_f}\mathbf{\hat e}_3 & \cdots
    \end{bmatrix}
    \label{eq:Acomponents}
\end{equation}
is the design matrix of $6\times4n$ that maps the input forces into the total force and torque of the structure.
In traditional  multi-rotor vehicles, this matrix is fixed after building the robot. In our case, this matrix can change depending on the module configuration.

Combining \eqref{eq:newton}, \eqref{eq:euler}, and \eqref{eq:MAu}, we obtain the mapping from input vector $\mathbf{u}$ to the linear and angular acceleration of the structure in the world reference frame, 
\begin{equation}
    \begin{bmatrix}
        \ddot{\mathbf{r}}\\
        \dot{\boldsymbol\omega}
    \end{bmatrix} = 
    \begin{bmatrix}
        \frac{1}{nm}{}^W\mathbf{R}_S & \mathbf{0}\\
        \mathbf{0} & \mathbf{I}_S^{-1}
    \end{bmatrix}\mathbf{A}\mathbf{u} - \begin{bmatrix}
        g\mathbf{\hat e}_3\\
        \mathbf{I}_S^{-1}(\boldsymbol\omega\times\mathbf{I}_S\boldsymbol\omega)
    \end{bmatrix}.
    \label{eq:dynamicsLinear}
\end{equation}
Note that the number of controllable DOF depends on the rank of $\mathbf{A}$.
For a quadrotor, the number of controllable DOF is four \cite{5717652}. 
A hexarotor with tilted propellers can have six controllable DOF, defining a fully actuated robot \cite{franchi2018full}.
We want to design modules with tilted propellers such that they fly by themselves, and when combined, they can have 4, 5, or 6 controllable DOF. We design a controller for each case and make sure the control strategies are general such that they work for any number of modules in a structure. The problems are stated as follows.
\begin{problem}[Module Design] Given the desired direction of maximum thrust of a module, specified by the rotation matrix $\mathbf{R}^\star$, find the tilting directions of its propellers, ${}^{M_i}\mathbf{R}_j$, such that when
all propellers are generating an identical thrust, the total thrust is parallel to the vector $\mathbf{R}^\star\mathbf{\hat e}_3$ and the total moment is zero.
\label{p:Rmoduledesign}
\end{problem}
\begin{problem}[Controlling H-ModQuad]
Given a structure with either 4, 5 or 6 controllable degrees of freedom, 
derive a controller that optimizes the use of its maximum force.
\label{p:control}
\end{problem}
Problem \ref{p:Rmoduledesign} is discussed in Section \ref{sec:modDesign} and Problem \ref{p:control} is explored in Section \ref{sec:control}.

\section{Module Design}
{
    \label{sec:modDesign}
    \noindent
    In order to develop a module that can fly and dock to other modules, we need the module to be balanced, 
    meaning that the module can hover without rotating while all propellers are generating identical forces.
    In this section, we describe the factors to design a balanced module.
    
    The matrix $\mathbf{A}$ is a $6\times4$ matrix for a single module. 
    The orientation of each propeller, ${}^{M_i}\mathbf{R}_{j}$, can be written as a function of two angles $\alpha_j, \beta_j \in [-\pi/2, \pi/2]$,
    \begin{equation}
        {}^{M_i}\mathbf{R}_{j} = \text{Rot}({y}, \beta_j)\,\text{Rot}({x}, \alpha_j),
        \label{eq:ori}
    \end{equation}
    where the function \textit{Rot} returns the rotation matrix of rotating the given angle along the specified axis.
    
    {
    In this formulation, Problem \ref{p:Rmoduledesign} is reduced to finding a combination of $\alpha_j$, $\beta_j$ for the four propellers and an unknown scalar value $\lambda>0$, such that when all the propellers generate a unit force, i.e., $\mathbf{u}=\mathbf{1}$, the force and torque of the module are
    $$\boldsymbol\tau=\mathbf{0}\text{ and } \mathbf{f}^\star=\lambda\mathbf{\hat{f}}^\star,$$
    where $\mathbf{\hat{f}}^\star=\mathbf{R}^\star\mathbf{\hat e}_3$ is a unit vector that determines where the thrust points at and $\lambda$ is the magnitude of the thrust force.
    The vector $\mathbf{\hat{f}}^\star$ is defined during the module design.}

    Based on \eqref{eq:taustructure}, the torque of the module 
    is generated by the force and drag of the propeller,
    i.e., $\boldsymbol\tau=\boldsymbol\tau_f+\boldsymbol\tau_d$. 
    Considering the fact that typically the torque created by the force is much larger than that by the drag, i.e., $k_f >> k_m$,
    a solution for $\boldsymbol\tau=\mathbf{0}$ can be obtained by solving $\boldsymbol\tau_f=\mathbf{0}$ \emph{and} $\boldsymbol\tau_d=\mathbf{0}$ independently. 
    %
    Therefore, from \eqref{eq:fstructure} and \eqref{eq:taustructure}, we obtain
    \begin{eqnarray}
        \textstyle\sum_j^4\mathbf{p}_j\times{}^{M_i}\mathbf{R}_{j}\mathbf{\hat e}_3   &=&\mathbf{0}\label{eq:Atf0},\\
        \textstyle\sum_j^4{}^{M_i}\mathbf{R}_{j}(-1)^j\mathbf{\hat e}_3&=&\mathbf{0}\label{eq:Atd0},\\
        \textstyle\sum_j^4{}^{M_i}\mathbf{R}_{j}\mathbf{\hat e}_3&=&{\lambda}\mathbf{R}^\star\mathbf{\hat e}_3.\label{eq:F}
    \label{eq:simplifiedMF}
    \end{eqnarray}
Therefore, every module that satisfies the constraints \eqref{eq:Atf0}-\eqref{eq:F} is balanced.
One solution that satisfies these constraints is making 
    $^{M_i}\mathbf{R}_{j}=\mathbf{R}^\star$ for all $j=1,...,4$.
    This leads us to define a specific type of module that creates a sub-group of balanced modules.
}

\begin{definition}[$R$-Module]
An \emph{$R$-module} is a module that is propelled by a quadrotor, and all its rotors are pointing in the same direction with respect to $\{M_i\}$, specified by rotation matrix ${}^{M_i}\mathbf{R}_j$.
\end{definition}
In this paper, we focus on the {$R$-module}, showing that it is balanced, however, there are other types of balanced modules that we leave for future studies.
\begin{proposition}
If a module is an R-Module, defined by the matrix $\mathbf{R}^\star$, the module is a balanced module.
\end{proposition}
\begin{proof}
By definition, ${}^{M_i}\mathbf{R}_j=\mathbf{R}^\star$.
In a quadrotor with a square configuration, the position of the propellers are $\mathbf{p}_1=-\mathbf{p}_3$ and $\mathbf{p}_1=-\mathbf{p}_3$, then, the torques are cancelled, satisfying~\eqref{eq:Atf0}.
Constraints \eqref{eq:Atd0} and \eqref{eq:F} are satisfied by replacing ${}^{M_i}\mathbf{R}_j=\mathbf{R}^\star$.
In \eqref{eq:F}, $\lambda=4$ for any $R$-Module with a non-identity $\mathbf{R}^\star$.
\end{proof}
A set of $R$-modules with different rotor orientations is called a \emph{heterogeneous team}. When the heterogeneous team forms a single rigid structure, it forms an \emph{H-Modquad structure}, or simply a \emph{heterogeneous structure}.
In the rest of this paper, the terms \emph{structure} and \emph{heterogeneous structure} are used interchangeably. By choosing different types of balanced modules to assemble the structure, one can configure and optimize the actuation properties of the structure to satisfy requirements of different tasks. 
\begin{figure*}[t!]
    \centering
    {\includegraphics[width=0.75\linewidth]{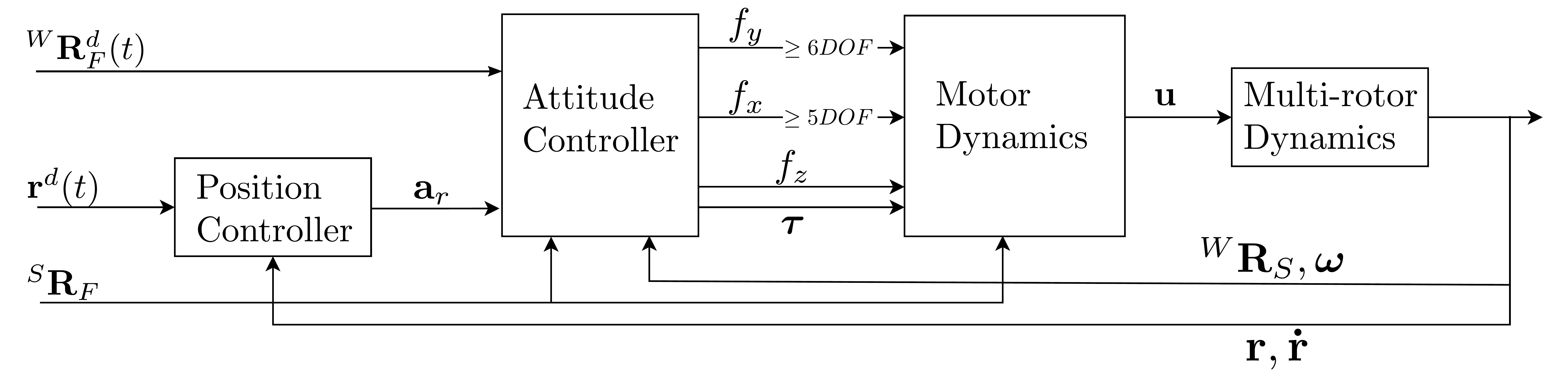}
    \caption{Control Diagram: The desired pose of $\{F\}$, $\mathbf{r}^d(t), {}^W\mathbf{R}^d_F(t)$ is tracked using the feedback on the pose of $\{S\}$, ${}^W\mathbf{R}_S(t), \mathbf{r}(t)$. The Motor Dynamics block is characterized by the matrices $\mathbf{A}_4^\dagger, \mathbf{A}_5^\dagger, \text{and }\mathbf{A}^\dagger$ depending on the number of controllable DOF. The rotation from $\{S\}$ to $F$-frame is specified by ${}^S\mathbf{R}_F$, which is fixed in $\{S\}$.}
    \label{fig:diagram}}
\end{figure*}

\section{Control}{
\label{sec:control}
\noindent
We separate the trajectory tracking control into two parts: position and attitude. {Fig. \ref{fig:diagram} shows an overview of the control strategy in a centralized way where all modules in the structure are controlled based on a trajectory function and a single stream of measurements. } 

Given a desired position, velocity, and acceleration, $\mathbf{r}^d$, $\mathbf{\dot r}^d$ and $\mathbf{\ddot r}^d$.
The position control determines the desired acceleration vector $\mathbf{a_r}$ by applying a PD controller on position error, $\mathbf{e_r} = \mathbf{r}^d-\mathbf{r}$, and velocity error, $\mathbf{e_{v}} = \mathbf{\dot r}^d-\mathbf{\dot r}\label{eq:ev}$, with a  feed-forward term,
\begin{equation}
        \mathbf{a_r} = \mathbf{K_re_r+K_ve_v}+g\mathbf{\hat e}_3+\mathbf{\ddot r}^d\label{eq:fd},
\end{equation}
where $\mathbf{K_r}$ and $\mathbf{K_v}$ are diagonal matrices with positive gains.

In fully-actuated vehicles, the position and the attitude control are independent. However, in under-actuated vehicles, e.g., a quadrotor, we need the attitude control to re-orient the vehicle to produce a linear acceleration that converges to~$\mathbf{a_r}$. Therefore, we need to take into account the controllable degrees of freedom to define the control strategies in $\mathsf{SE(3)}$. The controllable DOF depends on the matrix that maps rotor forces into the vehicle's force and torque. The matrix $\mathbf{A}$ in \eqref{eq:MAu} makes the map. Note in \eqref{eq:Acomponents} that $\mathbf{A}$ can be divided into two parts, $\mathbf{A}_{\mathbf{f}}$ and $\mathbf{A}_{\boldsymbol\tau}$ which are the first and last three rows of $\mathbf{A}$, respectively. Since a structure is composed of at least one module and the propellers of one module always generate linearly independent torques, the $\mathbf{A_\tau}$ portion of $\mathbf{A}$ in \eqref{eq:MAu} is full-row-rank, i.e., rank($\mathbf{A_\tau}$)=3. Docking more modules to the structure will add redundancy to the controllability of~$\boldsymbol\tau$. Hence, the rank of $\mathbf{A}$ depends on $\mathbf{A_f}$. The maximum total thrust that the structure can generate is not only related to the number of modules but also the orientation of its propellers. In order to maximize the thrust efficiency in the control policies, we introduce the concept of $F$-frame.
\begin{definition}[$F$-frame]
    The \emph{$F$-frame} of a structure, $\{F\}$, 
    is a coordinate frame with  origin is at the center of mass;
    the $z$-axis, $\mathbf{\hat z}_F$, points towards the direction where the structure can generate its maximum thrust;
    the $x$-axis, $\mathbf{\hat x}_F$, points towards the direction where the structure can generate its maximum thrust on the normal plane of $\mathbf{\hat z}_F$.
\end{definition}
%
In \cite{yoshikawa1985manipulability}, Yoshikawa introduced the concept of \emph{the manipulability ellipsoid} for robot-arm manipulation by
characterizing the response of a manipulator arm based on its Jacobian. 
We extend this concept for multi-rotor systems, characterizing the output force in response to the motor inputs.
Although we focus on forces to compensate the gravity, the same approach can be applied to map motor inputs to torques.
We determine the $F$-frame by using singular value decomposition on $\mathbf{A_f}$. 
For a given $\mathbf{A_f}$, we calculate the singular values and singular vectors of the rectangular matrix $\mathbf{A_f}$. 
Since the singular vectors are orthogonal, we use them to define the major and minor axes of the ellipsoid, transformed by $\mathbf{A_f}$ from a unit disc. We choose the singular vector associated with the largest singular value as the $z$-axis of $\{F\}$. In Fig. \ref{fig:overview}, we illustrate the ellipsoid for three different $2\times1$ structures projected on each  $xz$-plane.
\begin{figure}[b]
    \centering\includegraphics[clip, trim=0.cm 0.cm 0.cm .5cm,width=0.47\textwidth]{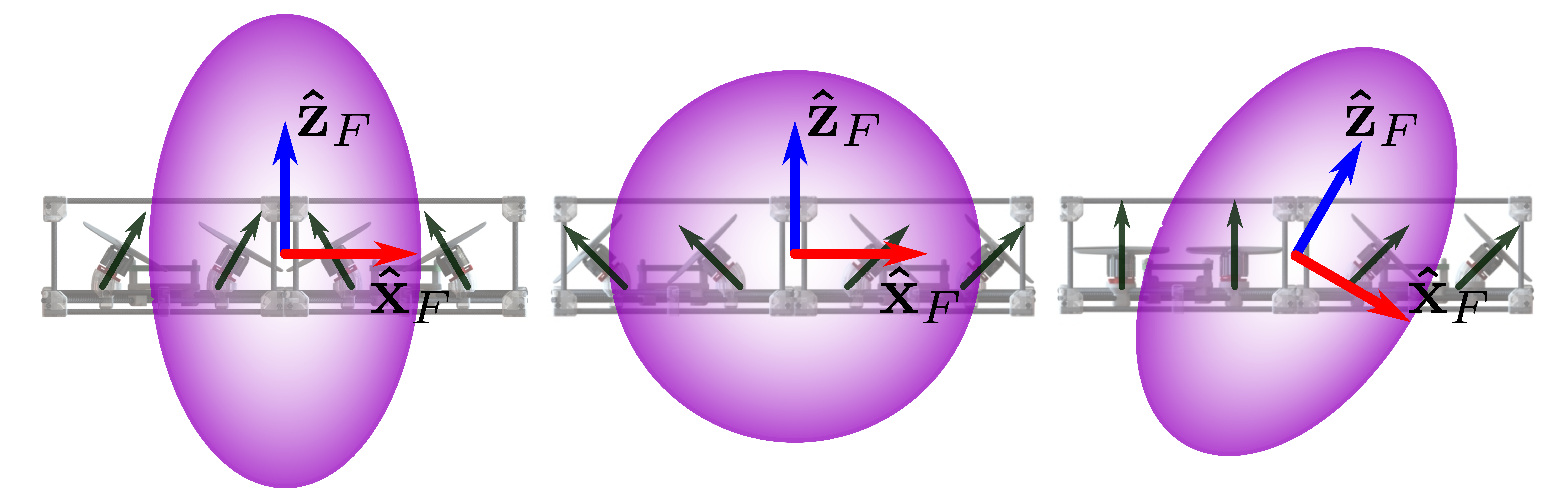}
    \caption{Side view of three H-ModQuad structures: each structure consists of two different modules. The black arrows represent the direction of rotor forces of the propellers. The purple ellipses represent the projection of the $F$-frame on the $xz$-plane of each structure. The first two structures have their $F$-frame aligned with $\{S\}$ while the last one has its $F$-frame deviating from $\{S\}$.
    }
    \label{fig:overview}
\end{figure}

In the case of $\text{rank}(\mathbf{A_f})=1$, 
all the force vectors from the rotors are linearly dependent in the same axis. We define the orientation of the axes of $\{F\}$ by rotating $\{S\}$ using same rotation matrix of the rotors in a module.
In the case of rank($\mathbf{A_f}$)>1, the singular vector associated with the second largest singular value is chosen as the $x$-axis of  $\{F\}$. The $y$-axis of $\{F\}$ is obtained by  $\mathbf{\hat y}_F=\mathbf{\hat z}_F\times\mathbf{\hat x}_F$. Then ${}^S\mathbf{R}_F=[\mathbf{\hat x}_F\; \mathbf{\hat y}_F\; \mathbf{\hat z}_F]$. 

We compute the attitude error in $\mathsf{SO(3)}$ with respect to the $F$-frame. Since the structure has attitude ${}^W\mathbf{R}_F$ and angular velocity $\boldsymbol{\omega}$, the angular tracking error is
\begin{eqnarray*}
        \mathbf{e_R} &=& \frac{1}{2}\left(\left({}^W\mathbf{R}_F^d\right)^{\top}{}^W\mathbf{R}_S{}^{S}\mathbf{R}_F-\left({}^W\mathbf{R}_S{}^{S}\mathbf{R}_F\right)^{\top}{}^{W}\mathbf{R}_F^d\right)^{\vee} \label{eq:eR},\\
        \mathbf{e}_{\boldsymbol\omega} &=& \boldsymbol\omega - \left({}^W\mathbf{R}_S{}^{S}\mathbf{R}_F\right)^{\top}{}^W\mathbf{R}_F^d\boldsymbol\omega^d, \label{eq:eomega}
    \label{eq:attitudeerror}
\end{eqnarray*}
where the ``\textit{vee}'' operator, $\vee$, maps a skew symmetric matrix to $\mathbb{R}^3$. 
{Note that by design, the output of the orientation sensors are with respect to the structure frame, $^W \mathbf{R}_S$. In order to maximize the actuation efficiency of the structure, our controllers track the attitude of $\{F\}$ to the desired attitude ${}^W\mathbf{R}_F^d$, which makes the attitude of $\{S\}$ to converge to ${}^W\mathbf{R}_F^d{}^S\mathbf{R}_F^\top$.}
Then, the necessary acceleration to track the attitude error is
\begin{eqnarray}
    \mathbf{a}_{\mathbf{R}} = -\mathbf{K_Re_R}-\mathbf{K_{\boldsymbol\omega} e_{\boldsymbol\omega}}\label{eq:tau4},    \label{eq:ftau4}
\end{eqnarray}
where $\mathbf{K_R}$ and $\mathbf{K}_{\boldsymbol\omega}$ are diagonal matrices with positive gains.
We can generate a torque to track that acceleration
$$\boldsymbol{\tau}=\mathbf{I}_S \mathbf{a_R} + \boldsymbol\omega\times \mathbf{I}_S\boldsymbol\omega.$$
We recall that different from \cite{5980409}, our attitude error is not with respect to the structure frame, $\{S\}$; we want to align $\{F\}$ with the desired attitude. 
Based on the concept of the $F$-frame, we proceed to define the control of the structure depending on the number of controllable DOF.

\subsection{Rank$(\mathbf{A})=4$}
\noindent
For a structure with rank$(\mathbf{A_f})=1$, similar to a traditional quadrotor, we can control the position and yaw angle of the structure using a geometric controller \cite{5717652}.
However, different from \cite{5717652}, we project the desired thrust vector on the $z$-axis of $\{F\}$, instead of $\{S\}$.
This approach reduces the energy consumption since the reference force is in the direction of the maximum force of the structure.
Given a desired acceleration vector $\mathbf{a}$ and a desired yaw angle $\psi^d$, the desired attitude ${}^W\mathbf{R}_F^d=[\mathbf{\hat x}^d\;\mathbf{\hat y}^d\;\mathbf{\hat z}^d]$ of the structure is obtained by
\begin{alignat}{3}
    \mathbf{\hat z}^d &= \frac{\mathbf{a_r}}{\Vert\mathbf{a_r}\Vert}
    , & & \qquad&\mathbf{\hat x}^c &= [\cos(\psi^d), \sin(\psi^d), 0]^{\top}%
    \nonumber
    \\
    \mathbf{\hat y}^d &= \frac{\mathbf{\hat z}^d\times\mathbf{\hat x}^c}{\Vert\mathbf{\hat z}^d\times\mathbf{\hat x}^c\Vert},
    & & \qquad&
    \mathbf{\hat x}^d &= \mathbf{\hat y}^d\times\mathbf{\hat z}^d. 
    \label{eq:direction4}
\end{alignat}
The thrust, $f$, is defined by the projection of $nm\,\mathbf{a_r}$ on $\mathbf{\hat z}_F$,
\begin{equation}
f = nm\,\mathbf{a_r}\cdot{}^W\mathbf{R}_S{}^{S}\mathbf{R}_F\mathbf{\hat e}\mathbf{}_3.\label{eq:f4}
\end{equation}
We can rewrite the motor dynamics, from \eqref{eq:MAu}, to use the four controllable DOF, 
$
    \begin{bmatrix}
        f \\ 
        \boldsymbol\tau
    \end{bmatrix}=\mathbf{A}_4\mathbf{u},
    \label{eq:A4}$
where $\mathbf{A}_4=
    \begin{bmatrix}
        \mathbf{v}^\top_{4f}\\
        \mathbf{A}_{\boldsymbol\tau}
    \end{bmatrix}$, and the vector $\mathbf{v}_{4f}\in\{-1,1\}^{4n}$ is associated to the direction of each propeller, either $1$ or $-1$  if propeller is pointing in the same or in the opposite direction of $\mathbf{\hat z}_F$, respectively. Since the matrix $\mathbf{A}_{4}$ is full-row rank, we can minimize the square sum of the propeller thrusts by applying the Moore-Penrose inverse
\begin{equation}
    \mathbf{u} = \mathbf{A}_4^\dagger\begin{bmatrix}
        f\\
        \boldsymbol\tau
    \end{bmatrix}.
    \label{eq:u4}
\end{equation}
The controller is exponentially stable and its proof follows the same logic in \cite{5980409} by using $^W\mathbf{R}_F$ instead of $^W\mathbf{R}_S$.
\subsection{Rank$(\mathbf{A})=5$}
\noindent
For a structure with at least two modules and rank($\mathbf{A_f}$)=$2$, the thrust vectors from all propellers are no longer co-linear but instead,  co-planar. Thus, the structure is able to control an additional DOF. Since the second major axis of $\{F\}$ is $\mathbf{\hat x}_F$, meaning that the structure is able to generate forces in $\mathbf{\hat x}_F$ independently, the additional DOF is given to the rotation around $\mathbf{\hat y}_F$. We choose to use this additional DOF to track the pitch angle of $\{F\}$, $\theta^{d}$. The desired attitude, $^W\mathbf{R}^d_F=[\mathbf{\hat x}^d\,\:\mathbf{\hat y}^d\,\:\mathbf{\hat z}^d]$ of the structure is obtained by
\begin{alignat}{3}
        \mathbf{\hat z}^c &= \frac{\mathbf{a_r}}{\Vert\mathbf{a_r}\Vert}
        , & & \qquad&\mathbf{\hat x}^d &= \text{Rot}\left(z, \psi^d\right)\:\text{Rot}\left(y, \theta^d\right)\: \mathbf{\hat e}_1,
        \nonumber\\
        \mathbf{\hat y}^d &= \frac{\mathbf{\hat z}^c\times\mathbf{\hat x}^d}{\Vert\mathbf{\hat z}^c\times\mathbf{\hat x}^d\Vert},
        & & \qquad& \mathbf{\hat z}^d&=\mathbf{\hat x}^d\times\mathbf{\hat y}^d.
    \label{eq:direction5}
\end{alignat}
Different from \eqref{eq:direction4}, in  \eqref{eq:direction5}, instead of projecting $\mathbf{\hat x}^c$ on the normal direction of $\mathbf{\hat y}^d \mathbf{\hat z}^d$-plane to obtain $\mathbf{\hat x}^d$, we obtain $\mathbf{\hat x}^d$ directly by applying the desired yaw and pitch angle on $\mathbf{\hat e}_1$. $\mathbf{\hat z}^c$ is projected on the normal direction of $\mathbf{\hat x}^d\mathbf{\hat y}^d$-plane to acquire the desired force in the $xz$-plane of $\{F\}$, where lies any possible force that the structure can generate. This method makes sure the desired attitude captures the desired pitch angle. The projected components of the desired force vector in the $xz$-plane of $\{F\}$ are
\begin{eqnarray*}
    f_z = nm\,\mathbf{a}_r\cdot\mathbf{\hat z}_F, \text{ and }
    f_x = nm\,\mathbf{a}_r\cdot\mathbf{\hat x}_F.\label{eq:f5}
\end{eqnarray*}
The columns of $\mathbf{A}_{\mathbf{f}}$ are co-planar vectors in the  $xz$-plane of $\{F\}$. Therefore, we remove the row in motor dynamics matrix $\mathbf{A}$, which relates $\mathbf{u}$ to the $\mathbf{\hat y}_F$-component of the resulted force vector, to obtain a $5\times4n$ matrix $\mathbf{A}_5$. Thus, we modify \eqref{eq:MAu} into
$
    \begin{bmatrix}
        f_z &
        f_x &
        \boldsymbol\tau^\top
    \end{bmatrix}^\top=\mathbf{A}_5\mathbf{u}.
    \label{eq:A5}
$
As long as $\mathbf{A}_5$ is full-row rank, we obtain the control input by computing the Moore-Penrose inverse,
\begin{equation}
\mathbf{u} = \mathbf{A}_5^\dagger\begin{bmatrix}
    f_z &
    f_x &
    \boldsymbol\tau^\top
\end{bmatrix}^\top.
\label{eq:u5}
\end{equation}
If $\mathbf{A}_5$ is not full-row rank, it means that all the rotors are in the $yz$-plane. So we can follow the same analyses but removing the row of associated with $f_x$ instead of $f_y$.

\subsection{Rank$(\mathbf{A})=6$}
\noindent
A structure with rank$(\mathbf{A})$=6 has rank($\mathbf{A}_{\mathbf{f}}$)=3. The structure is fully actuated and therefore, we can generate an acceleration vector $\mathbf{a}=[\mathbf{a_r}^\top, \mathbf{a_R}^\top]^\top$.
Based on \eqref{eq:dynamicsLinear}, the desired control input is 
\begin{equation}
    \mathbf{u} = \mathbf{A}^\dagger\begin{bmatrix}
        nm{}^W\mathbf{R}_S^\top & \mathbf{0}\\
        \mathbf{0} & \mathbf{I}_S
    \end{bmatrix}
    \left(
    \mathbf{a}
    + \begin{bmatrix}
        \mathbf{0}\\
        \boldsymbol\omega\times \mathbf{I}_S\boldsymbol\omega
    \end{bmatrix}\right).
    \label{eq:u6}
\end{equation}
\begin{figure}[t]
    \centering
    {\includegraphics[width=0.75\linewidth]{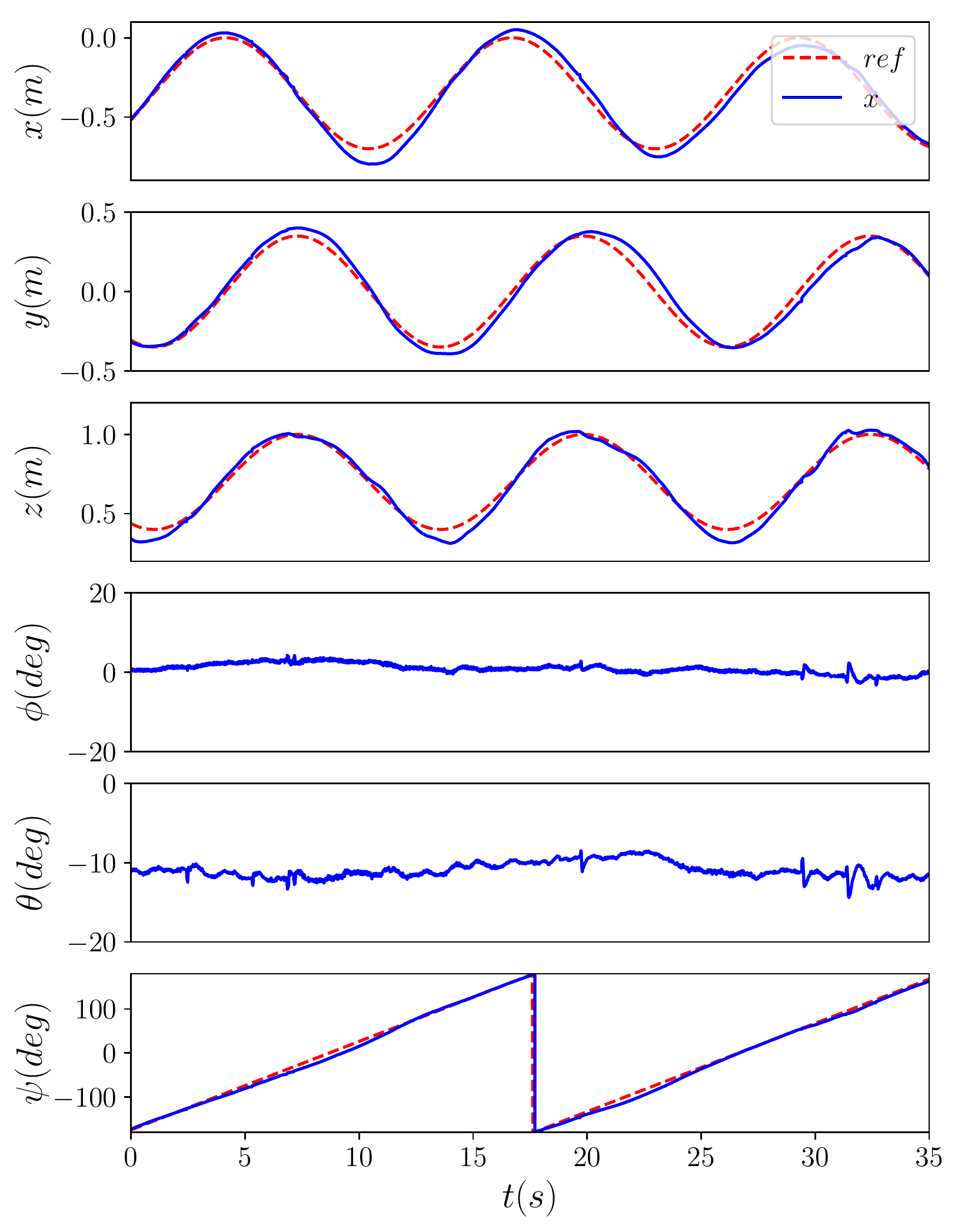}
    \caption{Experiment 1: An H-ModQuad structure of one module is tracking a helix. The dashed line represents the reference, and the solid line is the measurement. }    
    \label{fig:Exp1}}
\end{figure}
Due to motor saturation, the input vector $\mathbf{u}$ might not always feasible, which requires to limit the input $\mathbf{a}$. In the scope of this paper, we assume that the trajectory is feasible, and show the effectiveness of the control strategy when the desired pose is achievable.
}

\section{Experiments}{
\label{sec:exp}
\noindent
Our experimental testbed uses the Crazyflie-ROS framework \cite{crazyflieROS} with modified firmware to include the three controllers proposed in this paper. We use a motion capture system (Optitrack) operating at 120 Hz for vehicle localization. The modules measure their angular velocities using onboard IMU sensors; we store the predefined trajectories in the firmware as parametric functions of time. Time is synchronously updated using a radio transceiver.
For the controller implementation, we decentralize the attitude control and motor dynamics as in \cite{8461014}.



For each module, we use the Crazybolt board as a flight controller. The cuboid frame is composed of carbon fiber rods connected by 3-D printed vertices. Each brushless motor is mounted on a 3-D printed base with a cylindrical wedge. The angle of the wedge determines the motor orientation. Each module has a length of 12 cm, a height of 6 cm, and a weight of 135 g,  including a 520-mAh 2-cell LiPo battery. The modules dock rigidly by connecting their vertices using short rods. Fig. \ref{fig:titlePic} shows the prototype of four balanced modules and a structure composed of the four modules. 
We present three experiments on H-ModQuad structures with 4, 5, and 6 controllable DOF.
\begin{figure}[t]
\begin{center}$
\begin{array}{cc}
\includegraphics[clip, trim=0.1cm 0.4cm 0.cm 0.cm, width=0.2\textwidth]{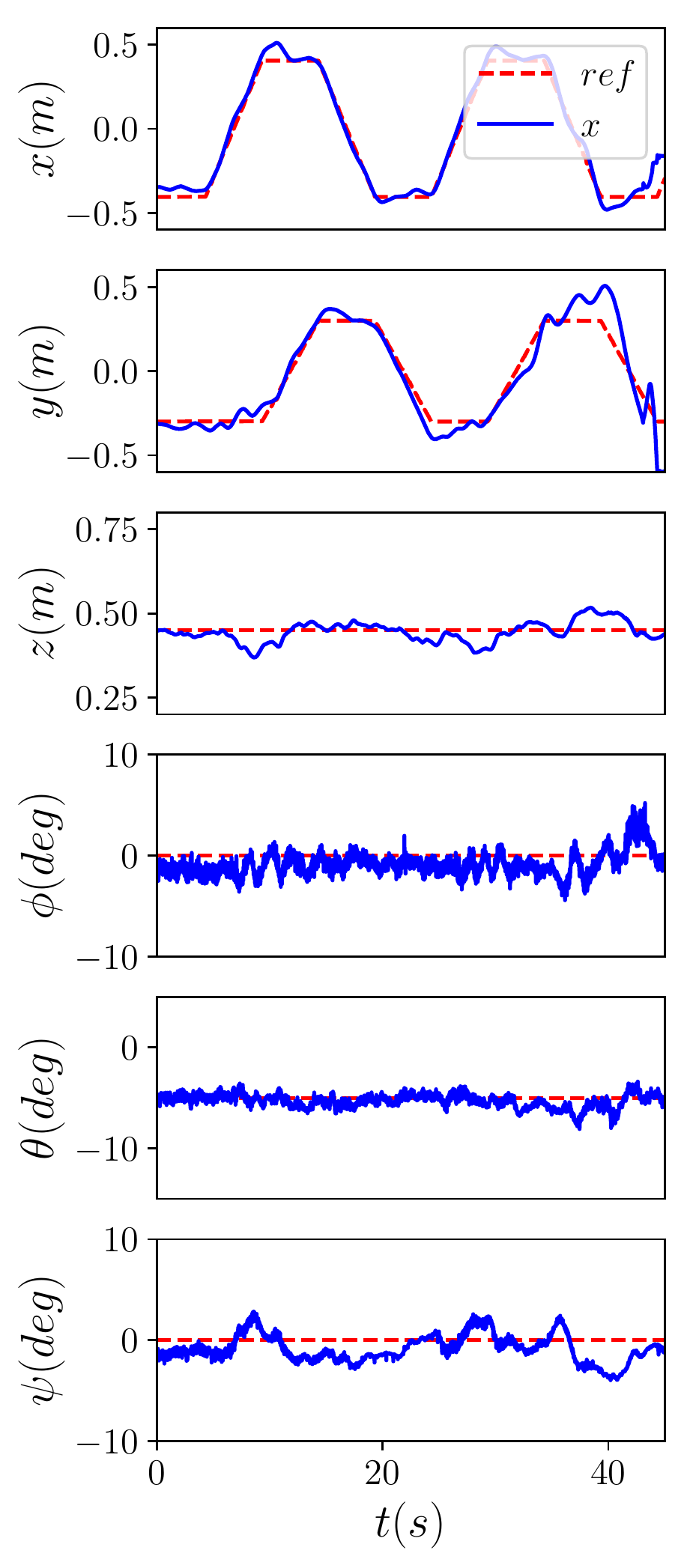}&
\includegraphics[clip, trim=0.1cm 0.4cm 0.cm 0.cm, width=0.2\textwidth]{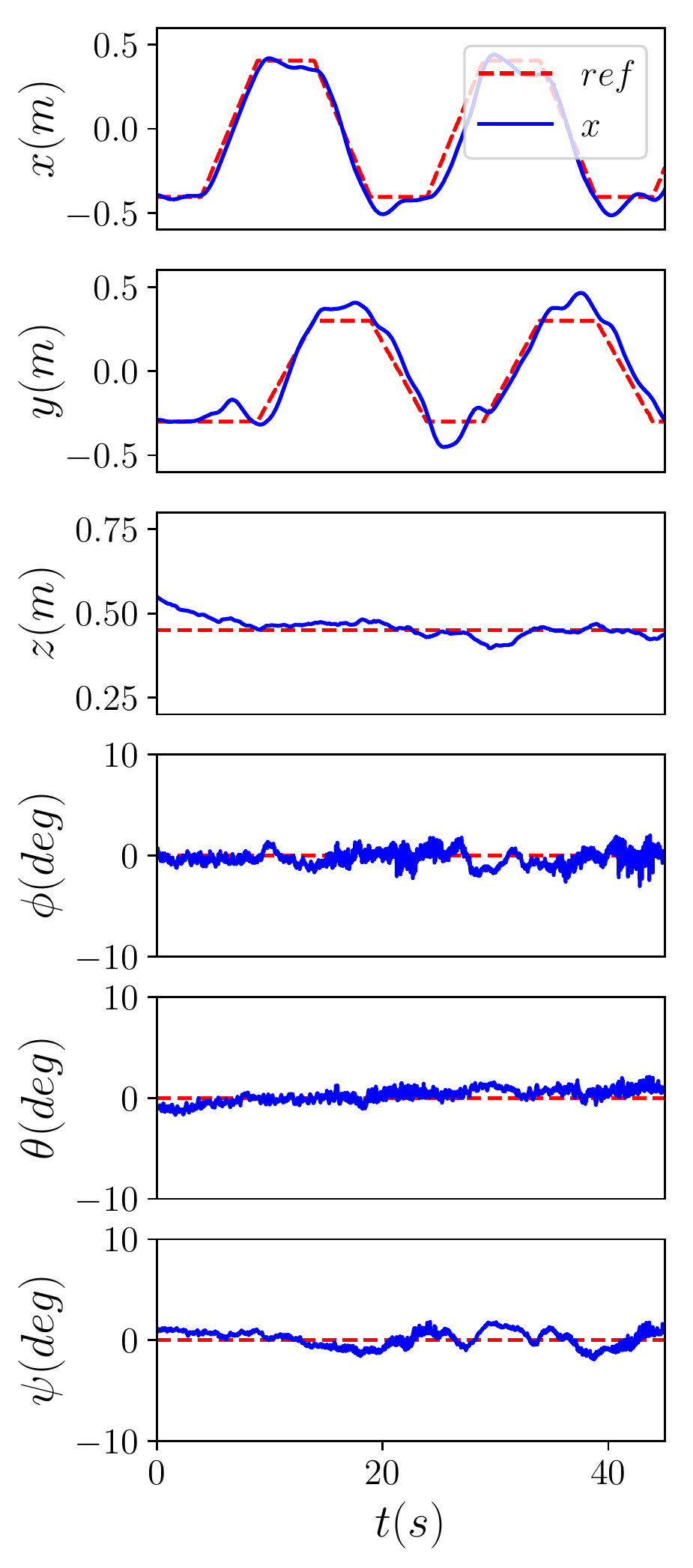}
\end{array}$
\end{center}
\caption{Experiment 2: An H-ModQuad structure of two modules track a rectangle with two \emph{different} fixed pitch angles. The plot on the left shows the structure following the rectangle while keeping a pitch angle of -5 degrees; on the right shows the structure tracking the rectangle while keeping a pitch angle of 0 degrees.} 
\label{fig:Exp2}
\end{figure}

\paragraph{4-DOF trajectory tracking}
We use an $R$-module with all its propellers tilting 10 degrees in pitch to track a 4-DOF trajectory (position and yaw orientation). Thus, the structure has the $F$-frame specified by ${}^S\mathbf{R}_F=\text{Rot}(y, \frac{\pi}{18})$.
We design a trajectory for the structure to move along a vertical helix in $\{W\}$. 
The helix centers at (-0.5, 0) in $xy$-plane with a radius of 0.45 and oscillates along the $z$-axis between 0.45 and 0.95 with a period of 14 seconds. 
While tracking the position, the structure is also rotating the yaw angle. 
The experiment result is shown in Fig.~\ref{fig:Exp1}. We highlight that the 4DOF structure of one module has an attitude ${}^W\mathbf{R}_S={}^S\mathbf{R}^\top_F$ during hovering, showing that we can drive the $F$-frame to track the desired attitude.
}

\paragraph{5-DOF rectangle tracking}{
\label{sec:exp2}
We use a 5-DOF structure, composed of two $R$-modules, to track a rectangle on the $xy$-plane of $\{W\}$ with a length of 0.8 m and a width of 0.6 m along $\mathbf{\hat y}_W$. One module has all its propellers tilting 30 degrees in pitch 
and the other 
tilting -30 degrees in pitch.
The resulted $F$-frame aligns with $\{S\}$, thus ${}^S\mathbf{R}_F$ is a $3\times3$ identity matrix. The structure can track the rectangle trajectory when its $\{S\}$ has a pitch angle of 0 \emph{and} $-5$ degrees, which illustrates the independence of its translation on $x$-axis and its pitch angle. The experiment result is shown in Fig. \ref{fig:Exp2}. We highlight when $\{S\}$ is pitching -$5$ degrees and moving in positive $x$-direction, the structure is moving in the opposite direction to where the structure can generate its maximum thrust.
}

\paragraph{6-DOF rectangle tracking}{

\begin{figure}[t]
    \centering
    {\includegraphics[width=0.75\linewidth]{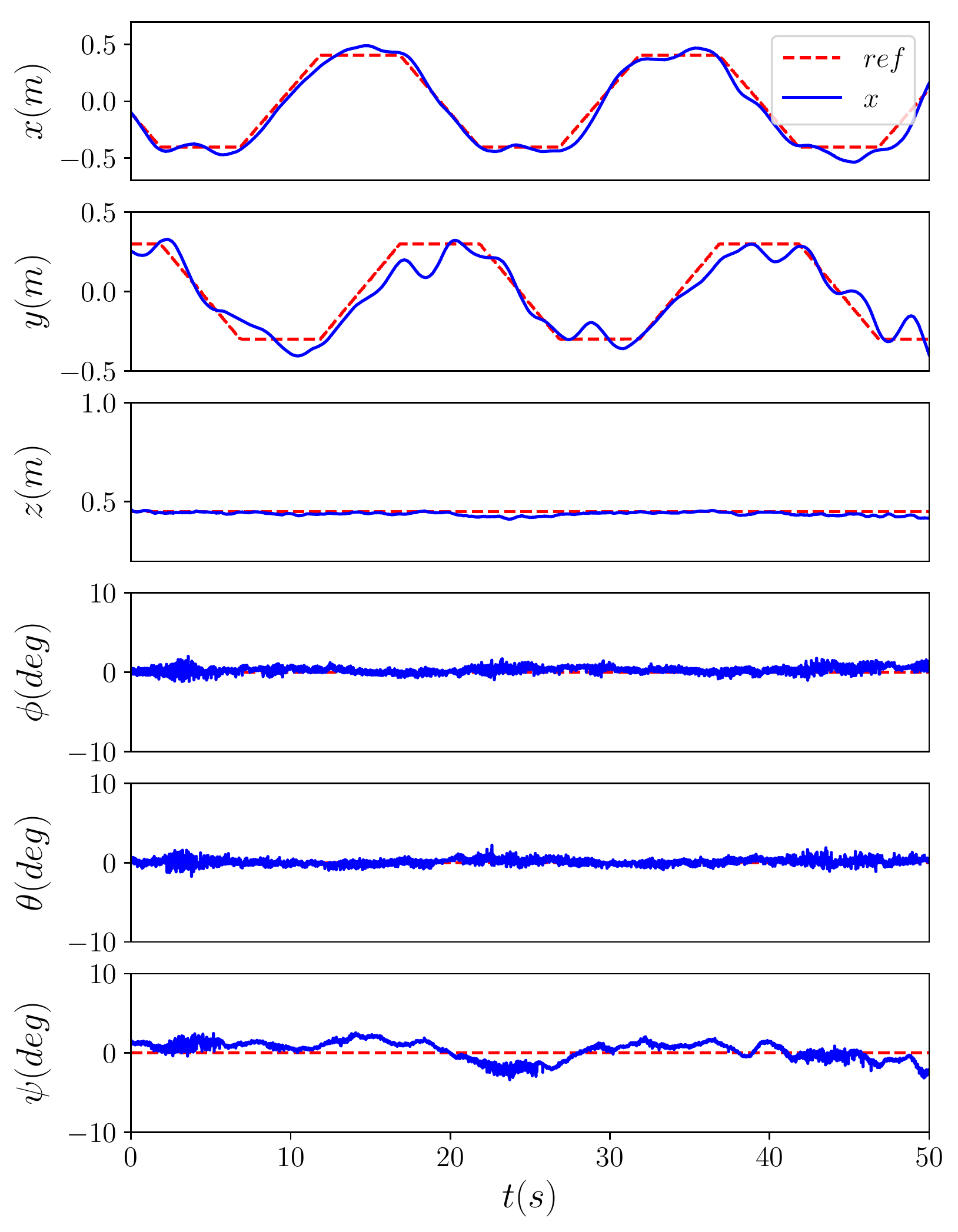}
    \caption{Experiment 3: An H-ModQuad structure of four H-ModQuad modules is tracking a rectangle with zero pitch and roll angles.}
    \label{fig:Exp3}}
\end{figure}

\noindent
We use a 6-DOF structure of four $R$-modules to track a rectangle with a fixed orientation. The modules have their propellers tilting $\pm$ 30 degrees in pitch, and $\pm$ 30 degrees in roll, respectively. The $F$-frame aligns with $\{S\}$, making ${}^S\mathbf{R}_F$ a $3\times3$ identity matrix. The experiment result is shown in Fig. \ref{fig:Exp3}. We highlight that when the structure is tracking the rectangle trajectory, its roll and pitch angles stay at 0 degrees, which shows the independence of translation and orientation. 
}


\section{Conclusion and Future Work}
\noindent
In this work, we propose a modular UAV that can increase its strength and controllable DOF from 4 to 5 and 6. 
Different types of modules define different orientations for their propellers. The combination of the heterogeneous modules determines the number of controllable degrees of freedom.
We extended the concept of the actuation ellipsoid to find the best reference orientation that can maximize the performance of the structure.
We propose three controllers, one for each number of controllable degrees of freedom.
The experiments in real robots show that we can control the degrees of freedom independently, i.e., the structure can translate independently on its orientation.

In future work, we want to explore the optimization of module configurations in a structure to satisfy characteristics required by given tasks. We also want to combine the three control policies introduced in this paper into a universal one and take into consideration the motor thrusting limits.


    
    

\bibliographystyle{IEEEtran}
\bibliography{ref}

\end{document}